\documentclass[runningheads]{llncs}
\usepackage{makeidx}
\usepackage{graphicx}
\usepackage{amsmath,amssymb}
\usepackage{rotating}
\usepackage{multirow}
\usepackage[lofdepth,lotdepth]{subfig}
\usepackage{tablefootnote}

\usepackage{color, colortbl}
\definecolor{Gray}{gray}{0.8}

\usepackage{hhline}

\usepackage{url}

\usepackage{color}

\newcommand{\eg}{e.g.,}
\newcommand{\ie}{i.e.,}
\newcommand{\etal}{et al.}

\newcommand{\vertex}{{\bf v}}
\newcommand{\pvertex}{{\bf p}}
\newcommand{\setC}{{\mathcal{C}}}
\newcommand{\setP}{{\mathcal{P}}}
\newcommand{\ld}{{\bf d}}
\newcommand{\lm}{{\bf m}}
\newcommand{\q}{{\bf q}}
\newcommand{\p}{{\bf p}}
\newcommand{\para}{{\bf \theta}}
\newcommand{\argmin}{\mathop{\mathgroup\symoperators argmin}}

\begin{document}
\pagestyle{headings}
\mainmatter

\title{A Comparison of Directional Distances for Hand Pose Estimation}

\titlerunning{A Comparison of Directional Distances for Hand Pose Estimation}
\authorrunning{D. Tzionas \and J. Gall}
\author{Dimitrios Tzionas \inst{1}\textsuperscript{,}\inst{2} \and Juergen Gall\inst{2}}
\urldef{\mailsa}\path|dimitris.tzionas@tuebingen.mpg.de|
\urldef{\mailsb}\path|gall@informatik.uni-bonn.de|
\institute{Perceiving Systems Department, MPI for Intelligent Systems, Germany\\ \mailsa \and Computer Vision Group, University of Bonn, Germany\\ \mailsb}

\maketitle

\setcounter{footnote}{0}

\begin{abstract}
	Benchmarking methods for 3d hand tracking is still an open problem due to the difficulty of acquiring ground truth data. 
	We introduce a new dataset and benchmarking protocol that is insensitive to the accumulative error of other protocols. 
	To this end, we create testing frame pairs of increasing difficulty and measure the pose estimation error separately for each of them. 
	This approach gives new insights and allows to accurately study the performance of each feature or method without employing a full tracking pipeline.
	Following this protocol, we evaluate various directional distances in the context of silhouette-based 3d hand tracking, expressed as special cases of a generalized Chamfer distance form.
	An appropriate parameter setup is proposed for each of them, and a comparative study reveals the best performing method in this context.
\end{abstract}

\vspace{-3mm}
\section{Introduction}
\vspace{-3mm}
\label{sec:intro}

	Benchmarking methods for 3d hand tracking has been identified in the review~\cite{Erol2007} as an open problem due to the difficulty of acquiring ground truth data. 
	As in one of the earliest works on markerless 3d hand tracking~\cite{570968}, quantitative evaluations are still mostly performed on synthetic data, \eg~\cite{RosalesASS01,Athitsos2003,1544833,LucaHands,Oikonomidis12}. 
	The vast majority of the related literature, however, is limited to visual, qualitative performance evaluation, where the estimated model is overlaid on the images. 
	
	While there are several datasets and evaluation protocols for benchmarking human pose estimation methods publicly available, 
	where markers~\cite{Dataset-HumanEva,AaLGTV11}, inertial sensors~\cite{BaaHel2010a}, or a semi-automatic annotation approach~\cite{Dataset_TUM_Kitchen} have been used to acquire ground truth data, 
	there are no datasets available for benchmarking articulated hand pose estimation. 
	We propose thus a benchmark dataset consisting of 4 sequences of two interacting hands captured by 8 cameras, where the ground truth position of the 3d joints has been manually annotated. 
	
	Tracking approaches are usually evaluated by providing the pose for the first frame and measuring the accumulative pose estimation error for all consecutive frames of the sequence, \eg~\cite{Dataset-HumanEva}. 
	While this protocol is optimal for comparing full tracking systems, it makes it difficult to analyze the impact of individual components of a system. 
	For instance, a method that estimates the joint positions with a high accuracy, but fails in a few cases and is unable to recover from errors, will have a high tracking error if an error occurs very early in a test sequence. 
	However, the tracking error will be very low if the error occurs at the end of the sequence. The accumulation of tracking errors makes it difficult to analyze in-depth situations where an approach works or fails. 
	We therefore propose a benchmark that analyzes the error not over a full sequence, but over a set of pairs consisting of a starting pose and a test frame. Based on the start pose and the test frame, the pairs have different grades of difficulty.     
	   
	In this work, we use the proposed benchmark to analyze various silhouette-based distance measures for hand pose estimation. 
	Distance measures that are based on a closest point distance, like the Chamfer distance, are commonly used due to its efficiency~\cite{570968} and often extended by including directional information~\cite{711175,1211346}. 
	Recently, a fast method that computes a directional Chamfer distance using a 3d distance tensor has been proposed~\cite{Fast_Chamfer} for shape matching. 
	In this work, we introduce a general form of the Chamfer distance for hand pose estimation and quantitatively compare several special cases.

	\begin{figure}[t]
	\centering
		\subfloat[subfigure 1 lalala][initial pose]{
			\includegraphics[width=0.29\textwidth]{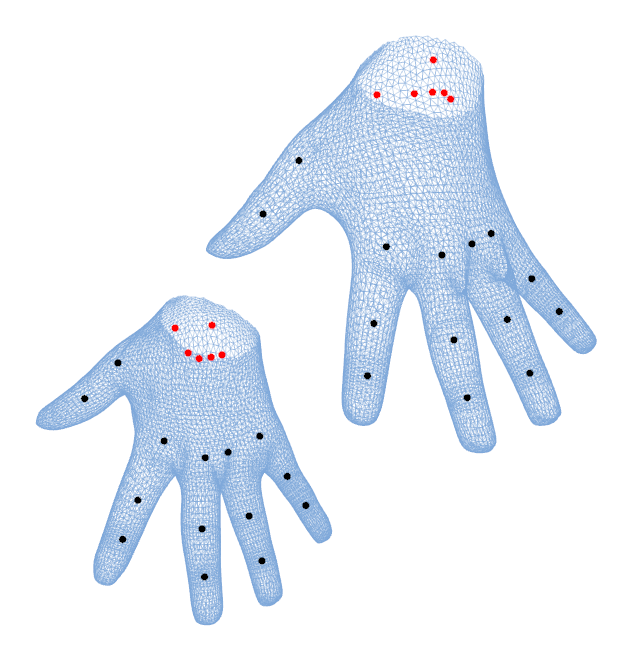}
			\label{fig:joints_Subfig1}
		}
		\quad
		\subfloat[subfigure 2 lalala][target silh. (synthetic)]{
			\includegraphics[width=0.29\textwidth]{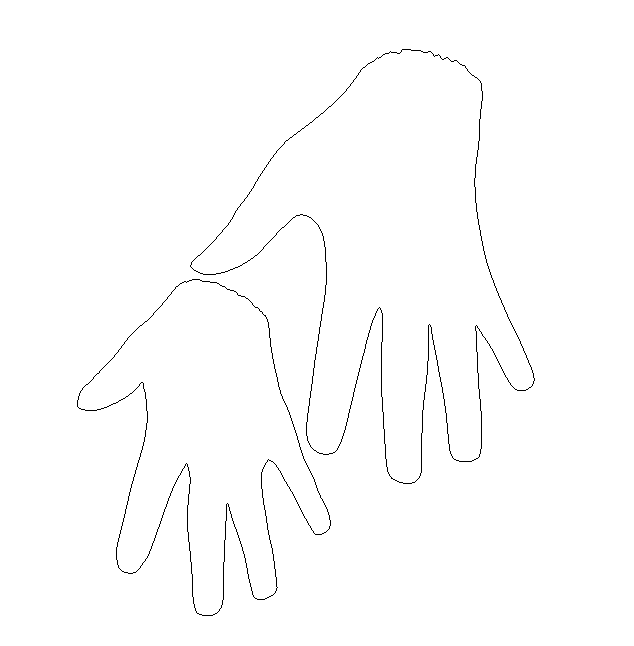}
			\label{fig:joints_Subfig2}
		}
		\quad
		\subfloat[subfigure 2 lalala][target silh. (realistic)]{
			\includegraphics[width=0.29\textwidth]{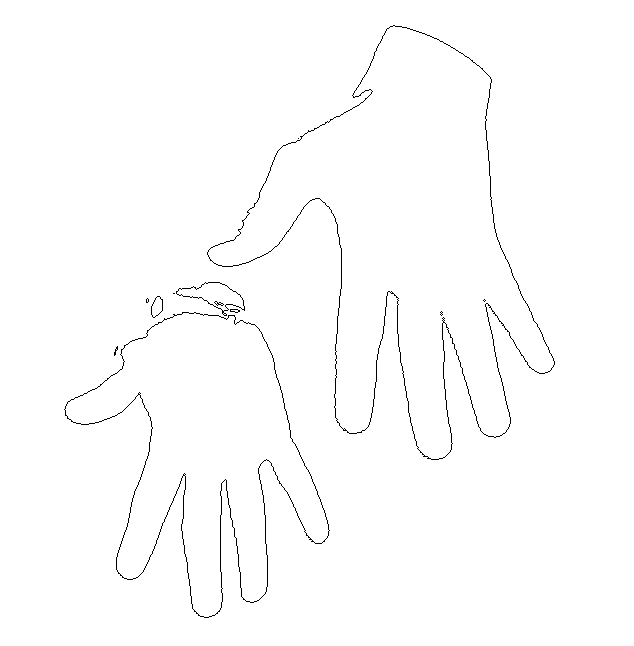}
			\label{fig:joints_Subfig2}
		}
	\caption{ 
		{\it Initial pose} (a) and synthetic (b) and realistic (b) {\it target silhouettes} of one camera view. 
		The benchmark measures the pose estimation error of the joints of both hands.
		In the synthetic experiments all joints (\emph{all} dots in (a)) are taken into account, while in the realistic only a subset (\emph{black} dots in (a)) is evaluated.
	}
	\label{fig:jointsAndStartingEnding}
	\vspace{-3mm}
	\end{figure}

\vspace{-1mm}
\section{Related Work}
\vspace{-3mm}
\label{sec:relwork}
	
	Since the earliest days of vision-based hand pose estimation~\cite{Rehg1994,Erol2007}, 
	low-level features like silhouettes~\cite{570968}, edges~\cite{Heap:19967}, depth~\cite{DelamarreF01}, optical flow~\cite{570968}, shading~\cite{GorceFP11} or a combination of them~\cite{LU03} have been used for hand pose estimation.
	Although Chamfer distances combined with an edge orientation term have been used in~\cite{1211346,Athitsos2003,Sudderth2004a,Stenger2006}, the different distances have not been thoroughly evaluated for hand pose estimation. 
	While a KD-tree is used in~\cite{Sudderth2004a} to compute a directional Chamfer distance, Liu \etal~\cite{Fast_Chamfer} recently proposed a distance transform approach to efficiently use a directional Chamfer distance for shape matching.   
	Different methods of shape matching for pose estimation have been compared in the context of rigid objects~\cite{Compare_CM_ICP} or articulated objects~\cite{Compare_LCS_String}. 
	While previous work mainly considered to estimate the pose of a hand in isolation, recent works consider more complicated scenarios where two hands interact with each other~\cite{Oikonomidis12,LucaHands} or with objects~\cite{Hamer2009,RomeroKK10,Hamer2010,Oikonomidis11,LucaHands}.

\vspace{-1mm}
\section{Hand Pose Estimation}
\vspace{-3mm}
\label{sec:handpose}
	
	For evaluation, we use a publicly available hand model~\cite{LucaHands}, 
	consisting of a set of vertices, an underlying kinematic skeleton with 35 degrees of freedom (DOF) per hand, and skinning weights. The vertices and the joints of the skeleton are shown in Fig.~\ref{fig:jointsAndStartingEnding}.
	Each 3d vertex $\vertex$ is associated to a bone $j$ by the skinning weights $\alpha_{\vertex,j}$, where $\sum_j \alpha_{\vertex,j} = 1$.
	The articulated deformations of a skeleton are encoded by the vector $\para$ that represents the rigid bone transformations $T_j(\para)$, \ie\ rotation and translation, by twists ${\hat{\xi}\in se(3)}$~\cite{Murray_MathInRob,Malik_Twist}. 
	Each twist-encoded rigid body transformation $\para_j\hat{\xi}_j$ for a bone $j$ can be converted into a homogeneous transformation matrix by the exponential map operator, \ie\ ${T_j(\para) = \exp(\theta_j \hat{\xi}_j)\in SE(3)}$. 
	The mesh deformations based on the pose parameters $\para$ are obtained by the linear blend skinning operator~\cite{LBS_PoseSpace} using homogeneous coordinates:
	\begin{equation}\label{eq:lbs}
	\vspace{-1mm}
	\vertex(\para) = \sum_j \alpha_{\vertex,j} T_j(\para) \vertex\; .
	\vspace{-3mm}
	\end{equation}
	
	In order to estimate the hand pose for a given frame, correspondences between the mesh and the image of each camera $c$ are established. Each correspondence $(\vertex_{i},\q_i,c_i)$ associates a vertex $\vertex_{i}$ to a 2d point $\q_i$ in camera view $c_i$. 
	Assuming that the cameras are calibrated, the point $\q_i$ can be converted into a projection ray that is represented by the direction $\ld_i$ and moment $\lm_i$ of the line~\cite{STOL91,Rosenh_Plucker_IJCV}. 
	The hand pose can then be determined by the pose parameters that minimize the shortest distance between the 3d vertices $\vertex_i$ and 3d projection rays $(\ld_i,\lm_i)$:   
	\begin{equation}\label{eq:minpose}
	\vspace{-1mm}
	\argmin_{\para} \frac{1}{2N} \sum_{i=1}^N \Vert \vertex_{i}(\para) \times \ld_i - \lm_i \Vert^2\; .
	\vspace{-1mm}
	\end{equation}
	This non-linear least-squares problem can be iteratively solved~\cite{Rosenh_Plucker_IJCV}:
	\begin{itemize}
	 \item Extract correspondences for all cameras $(\vertex_{i},\q_i,c_i)\;$,
	 \item Solve \eqref{eq:minpose} using the linearization $T_j(\para) = \exp(\theta_j \hat{\xi}_j) \approx I + \theta_j\hat{\xi}_j\;$,
	 \item Update vertex positions by \eqref{eq:lbs}.
	\end{itemize}
	In this work, we reformulate \eqref{eq:minpose} as a Chamfer distance minimization problem.

\vspace{-1mm}
\section{Generalized Chamfer Distance}
\vspace{-3mm}
\label{sec:chamfer}

	As discussed in Section~\ref{sec:relwork}, the Chamfer distance is commonly used for shape matching and has been also used for pose estimation by shape matching. 
	In our context, the Chamfer distance between pixels of a contour $\setC$ for a given camera view and the set of projected rim vertices $\setP(\para)$, 
	which depend on the pose parameters $\para$ and project onto the contour of the projected surface, is
	\begin{equation}\label{eq:chamfer}
	\vspace{-1mm}
	d_{Chamfer}(\para,\setC) = \frac{1}{\vert \setP(\para) \vert} \sum_{\pvertex \in \setP(\para)} \min_{\q \in \setC} \Vert \pvertex - \q \Vert \; .
	\vspace{-1mm}
	\end{equation}
	This expression can be efficiently computed using a 2d distance transform~\cite{DT_Felzenszwalb}.

\newpage
	The Chamfer distance \eqref{eq:chamfer} can be generalized by 
	\begin{equation}\label{eq:chamfergeneral}
	\vspace{-1mm}
	d_{Chamfer}^{Z,f,d}(\para,\setC) = \frac{1}{Z} \sum_{\pvertex \in \setP(\para)} f\left( \p, \argmin_{\q \in \setC} d(\pvertex, \q) \right) \; ,
	\vspace{-1mm}
	\end{equation}
	\noindent where $d(\p,\q)$ is a 2d distance function to compute the distance between two points, $f(\p,\q)$ is a penalty function for two closest points, and $Z$ is a normalization factor.
	If we use
	\begin{equation}\label{eq:chamfer2}
	d(\p,\q) = \Vert \p - \q \Vert\; , \quad f(\p,\q) = d(\p,\q) \; , \quad Z = \vert \setP(\para) \vert\; ,
	\end{equation}
	$d_{Chamfer}^{Z,f,d}(\para,\setC)$ is the standard Chamfer distance \eqref{eq:chamfer}. 	
	In order to increase the robustness to outliers, $f(\p,\q) = \min\left( d(\p,\q)^2 , K \right)$ 
	is used in~\cite{Stenger2006}, where $K$ is a threshold on the maximum squared distance.    
	
	Orientation can be integrated by penalizing correspondences with inconsistent orientations:  
	\begin{equation}\label{eq:rejectcorr}
	d(\p,\q) = \Vert \p - \q \Vert\; , \quad 
	f(\p,\q) = 
	\begin{cases} 
	d(\p,\q) &\text{if} \quad \vert \phi(\p) - \phi(\q) \vert_{\phi} < \tau \\
	K & \text{otherwise}
	\end{cases}\; ,
	\quad Z = \vert \setP(\para) \vert\; ,
	\end{equation}
	or by computing the closest distance to points of similar orientation based on a circular distance threshold $\tau$~\cite{711175}:
	\begin{equation}\label{eq:quantize} 
	d(\p,\q) = 
	\begin{cases} 
	\Vert \p - \q \Vert &\text{if} \quad \vert \phi(\p) - \phi(\q) \vert_{\phi} < \tau \\
	\infty & \text{otherwise}
	\end{cases}\; , \quad f(\p,\q) = d(\p,\q) \; , \quad Z = \vert \setP(\para) \vert\; ,
	\end{equation}
	where $\vert \phi(\p) - \phi(\q) \vert_{\phi}$ is the circular distance between two angles, which can be signed, \ie\ in the range of $[0, \pi]$, or unsigned, \ie\ in the range of $[0, \frac{\pi}{2}]$.

	The directional Chamfer distance~\cite{Fast_Chamfer} can be written as
	\begin{equation}\label{eq:dt3}
	d(\p,\q) = \Vert \p - \q \Vert + \lambda \vert \phi(\p) - \phi(\q) \vert_{\phi}\; , \quad f(\p,\q) = d(\p,\q)\; , \quad Z = \vert \setP(\para) \vert \; .
	\end{equation} 

	To compute $d_{Chamfer}^{Z,f,d}(\para,\setC)$ with \eqref{eq:dt3} efficiently, $\phi$ can be quantized to compute a 3d distance transform~\cite{Fast_Chamfer}. As in~\cite{Fast_Chamfer}, we compute $\phi(\q)$ by converting $\setC$ into a line representation~\cite{approxContour}. 
	$\phi(\p)$ is obtained by projecting the normals of the corresponding vertices in $\setP(\para)$.  

	In order to use the generalized Chamfer distance $d_{Chamfer}^{Z,f,d}(\para,\setC)$ for pose estimation from multiple views \eqref{eq:minpose}, only $f$ and $Z$ need to be adapted. 
	Let $\setC(c)$ denote the contour of camera view $c$ and $\setP(\para,c)$ the set of projected vertices for pose parameters $\para$ and camera $c$. \eqref{eq:minpose} can be rewritten as 
	\begin{align}\label{eq:minposecm}
	&\argmin_{\para} \frac{1}{2 \sum_c \vert \setP(\para,c) \vert }\sum_c d_{Chamfer}^{Z,f,d}(\para,\setC(c)) \\ 
	\text{with}\qquad &f(\p,\q) = \Vert \vertex(\para) \times \ld - \lm \Vert^2 \; , Z = 1 \; ,
	\end{align}
	where $\vertex(\para)$ is the 3d vertex corresponding to $\p \in \setP(\para)$ and $(\ld,\lm)$ is the 3d projection ray corresponding to $\q$. 
	$d(\p,\q)$ can be any of the functions \eqref{eq:chamfer2}-\eqref{eq:dt3}.
	
	In case of \eqref{eq:rejectcorr}, instead of adding a fixed penalty term $K$, correspondences with inconsistent orientation can be simply removed and $\setP(\para,c)$ becomes the set of correspondences with $\vert \phi(\p) - \phi(\q) \vert_{\phi} < \tau$.

\vspace{-1mm}
\section{Benchmark}\label{sec:bench}
\vspace{-3mm}
	We propose a benchmarking protocol that analyzes the error not over full sequences, but over a sampled {\it set of testing pairs}. Each pair consists of a {\it starting pose} and a {\it test frame}, ignoring the intermediate frames to simulate various difficulties.
	This approach gives new insights and provides means to analyze in-depth the contributions of various features or methods to the overall tracking pipeline under varying difficulty and to thoroughly study failure cases.
	
	In this respect, 4 publicly available sequences\footnote{Model, videos, and motion data are provided at \url{http://cvg.ethz.ch/research/ih-mocap}. Sequences: 
	\emph{Finger tips touching and praying, Fingers crossing and twisting, Fingers folding, Fingers walking}. 
	Video: $1080\times1920$ px, 50 fps, 8 camera-views. 
	} are used, containing realistic scenarios of two strongly interacting hands~\cite{LucaHands}.
	$10\%$ of the total frames are randomly selected, forming the set of {\it test frames} of the final pairs. 
	This is the basis to create 4 different sets of image pairs, having 1,5,10,15 frames difference respectively 
	between the {\it starting pose} and the {\it test frame}, presenting thus increasing difficulty for tracking systems. 
	These 4 sets and the overall combination constitute a challenging dataset, representing realistic scenarios the occur due to low frame rates, fast motion or estimation errors in the previous frame.
	
	The created testing sets are used in two experimental setups: a purely {\it synthetic} and a {\it realistic}. 
	In both cases, the {\it starting pose} is given by the publicly available motion data outputed by the tracker of~\cite{LucaHands}. 
	In the {\it synthetic} experimental setup the {\it test frame} is synthesized by the hand model and the aforementioned motion data, while the required ground truth exists inherently in them.
	In the {\it realistic} setup the {\it test frame} is given by the camera images, for which no ground-truth data are available, thus the frames have been manually annotated\footnote{The ground-truth annotated dataset, along with a viewer-application,  is available at \url{http://files.is.tue.mpg.de/dtzionas/GCPR_2013.html}.}. 
	As error measure, we use the average of the Euclidean distances between the estimated and the ground-truth 3d positions of the joints.
	For the realistic setup we use only the joints of the model that could be annotated, which are depicted with black color in Fig.~\ref{fig:jointsAndStartingEnding}. 
	For the synthetic setup all joints of the model (black and red) are taken into account.

\vspace{-2mm}
    \section{Experiments}
\vspace{-3mm}

	\subsection{Implementation Details}\label{sec:Experiments_ImplementationDetails}
	
		\vspace{-2mm}
				
		\begin{figure}[t]
		\centering
			\subfloat[subfigure 1 lalala][Synthetic]{
				\includegraphics[width=0.46\textwidth]{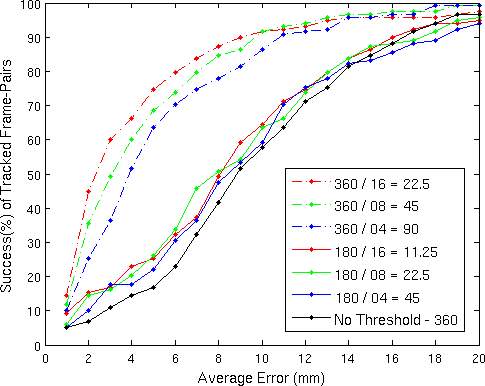}
				\label{fig:fig1_Subfig1}
			}
			\quad
			\subfloat[subfigure 2 lalala][Real]{
				\includegraphics[width=0.46\textwidth]{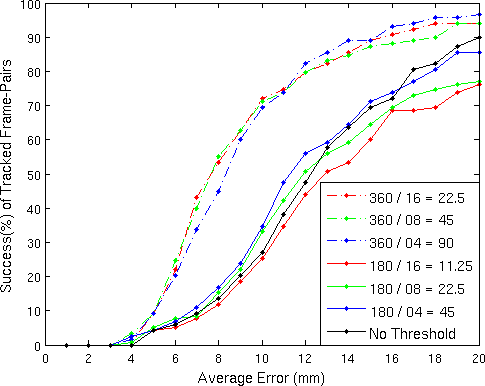}
				\label{fig:fig1_Subfig2}
			}
		\caption{
			Performance evaluation of \textbf{DCH-Thres} with different values of $\tau$ and both signed (360) and unsigned (180) distance $\vert \cdot \vert_{\phi}$. 
			The plots show the percentage of frame pairs (\textit{y-axis}) below a  given average error (\textit{x-axis}). 
			The {\it{signed}} distance (360) significantly outperforms the {\it{unsigned}} distance (180), 
			and the best performing circular distance threshold value is $\tau=22.5$.
		}
		\label{fig:fig1}
		\vspace{-4mm}
		\end{figure}
			
		The aforementioned benchmark is used to evaluate four special cases of the generalized Chamfer distance (Section~\ref{sec:chamfer}) for hand pose estimation.
	
		\textbf{CH} denotes the Chamfer distance without any orientation information~\eqref{eq:chamfer2}.
			
		\textbf{DCH-Thres} rejects correspondences if the orientations are inconsistent, depending on the circular distance threshold $\tau$~\eqref{eq:rejectcorr}.   
			
		\textbf{DCH-Quant} computes a 2d distance field for all quantizations of $\phi$ and assigns a vertex to one bin based on the orientation of its normal \eqref{eq:quantize}.  
		Instead of hard binning, soft binning can also be performed, denoted by \textbf{DCH-Quant2}. 
		In this case, the two closest bins are used, yielding two correspondences per vertex.
		
		\textbf{DCH-DT3} denotes the approximation of the directional Chamfer distance~\eqref{eq:dt3} proposed by Liu \etal~\cite{Fast_Chamfer}.
		The approach computes a 3d distance field \emph{DT3} and depends on two parameters.
		While $\lambda$ steers the impact of the orientation term in~\eqref{eq:dt3}, $\phi$ is quantized by a fixed number of bins.
		
		As mentioned in Section~\ref{sec:chamfer}, the {\it{target silhouette}} is approximated with linear line segments for all the directional distances \emph{DCH}, using~\cite{approxContour}.
		We also investigate two versions of the circular distance $\vert \cdot \vert_{\phi}$, namely the unsigned version, denoted by \emph{180}, and the signed version, denoted by \emph{360}.   	
			
\vspace{-4mm}
	\subsection{Results}
	
	\vspace{-2mm}
		
		\begin{figure}[t]
		\centering
			\subfloat[subfigure 1 lalala][Synthetic]{
				\includegraphics[width=0.46\textwidth]{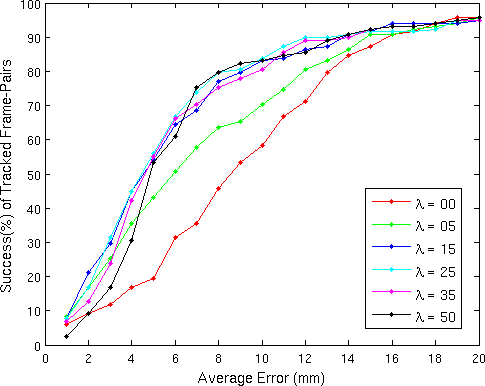}
				\label{fig:fig2_Subfig1}
			}
			\quad
			\subfloat[subfigure 2 lalala][Real]{
				\includegraphics[width=0.46\textwidth]{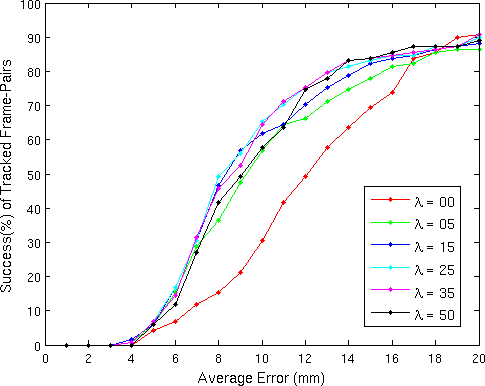}
				\label{fig:fig2_Subfig2}
			}
			\quad
			\subfloat[subfigure 3 lalala][Synthetic]{
				\includegraphics[width=0.46\textwidth]{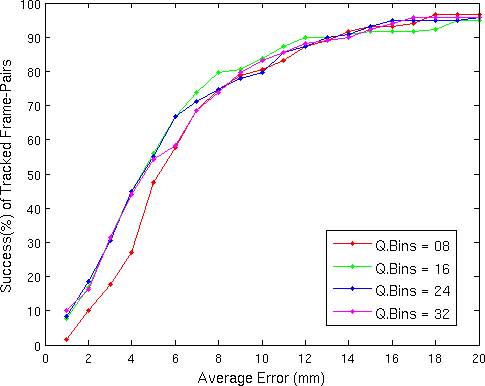}
				\label{fig:fig2_Subfig3}
			}
			\quad
			\subfloat[subfigure 4 lalala][Real]{
				\includegraphics[width=0.46\textwidth]{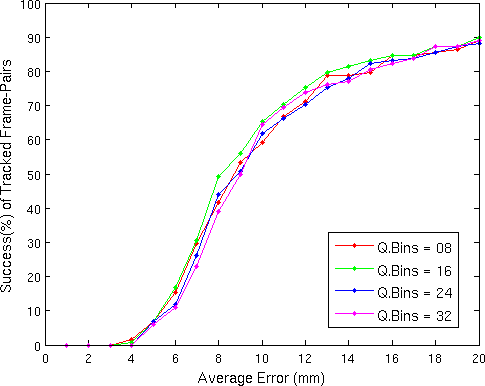}
				\label{fig:fig2_Subfig4}
			}
		\caption{
			\textbf{(a-b)} Performance evaluation of \textbf{DCH-DT3} with different values of $\lambda$, using 16 quantization bins. 
			While the orientation term significantly improves the performance, the performance gets saturated for values in the range [15,35]. 
			\textbf{(c-d)} Performance evaluation of \textbf{DCH-DT3} with different quantizations of $\phi$, using $\lambda = 25$. 
			The synthetic data shows that more than 8 bins are required, though the differences are rather small on the real dataset.
			This is in accordance with Fig.~\ref{fig:fig1} since a threshold of $22.5$ corresponds to 16 quantization bins.                 
		}
		\label{fig:fig2}
		\vspace{-5mm}
		\end{figure}
			
		We have evaluated all Chamfer distances both on the synthetic and the realistic dataset in order to compare the distances for 3d hand pose estimation, 
		but also in order to investigate the performance predicting abilities of synthetic test data.  
		As measure, we use the average joint error per test frame and compute the percentage of frames with an error below a given threshold. 
		We first evaluated the differences between the signed and unsigned circular distance for \textit{DCH-Thres} and varied the threshold parameter $\tau$. 
		The results are plotted in Fig.~\ref{fig:fig1}. The plot shows that the signed distance outperforms the unsigned distance. 
		Since we observed the same result for \textit{DCH-DT3}, we only report results for the signed distance (360) in the remaining experiments.  
		
		For \textit{DCH-DT3}, we evaluated the impact of the two parameters $\lambda$ and the number of quantization bins for $\phi$.   
		The results are plotted in Fig. \ref{fig:fig2}.
		Figs.~\ref{fig:fig2_Subfig1} and~\ref{fig:fig2_Subfig2} show the importance of directional information for hand pose estimation, 
		and reveal that there is a large range of $\lambda$ that works well. 
		With a finer quantization of $\phi$, the original directional Chamfer distance \eqref{eq:dt3} is better approximated. 
		Figs.~\ref{fig:fig2_Subfig3} and~\ref{fig:fig2_Subfig4} show that 16 bins are sufficient for this task. 
					
\begin{figure}[t]
		\centering
			\subfloat[subfigure 1 lalala][Synthetic]{
				\includegraphics[width=0.46\textwidth]{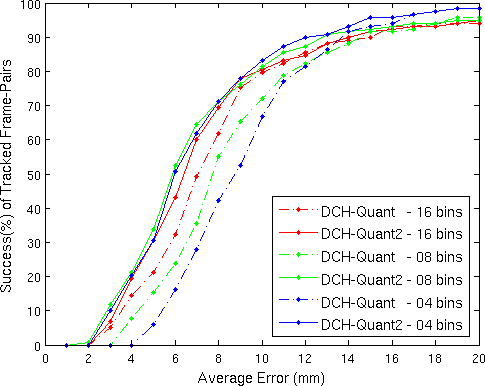}
				\label{fig:fig4_Subfig1}
			}
			\quad
			\subfloat[subfigure 2 lalala][Real]{
				\includegraphics[width=0.46\textwidth]{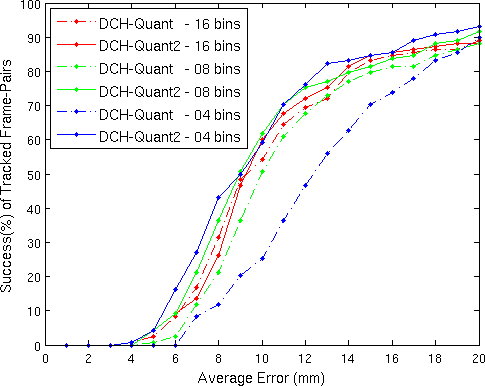}
				\label{fig:fig4_Subfig2}
			}
		\vspace{-2mm}
		\caption{
			Performance evaluation of \textbf{DCH-Quant} and \textbf{DCH-Quant2} with different quantizations of $\phi$. 
			Soft-binning outperforms hard assignments and in this case fewer bins perform better than many bins.
		}
		\label{fig:fig4}
		\vspace{-2mm}
		\end{figure}

		We finally evaluated the number of bins for \textit{DCH-Quant} and \textit{DCH-Quant2}. 
		Fig.~\ref{fig:fig4} shows that \textit{DCH-Quant2} performs better than \textit{DCH-Quant}. 
		In this case, a large number of bins results in a very orientation sensitive measure, and the performance decreases with a finer quantization, in contrast to \textit{DCH-DT3}.  
		
		\begin{figure}[]
		\centering
			\subfloat[subfigure 1 lalala][Synthetic]{
				\includegraphics[width=0.46\textwidth]{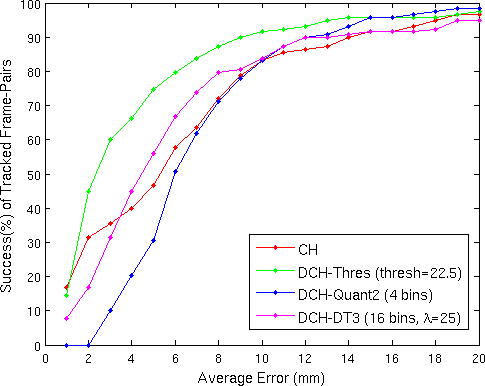}
				\label{fig:fig5_Subfig1}
			}
			\quad
			\subfloat[subfigure 2 lalala][Real]{
				\includegraphics[width=0.46\textwidth]{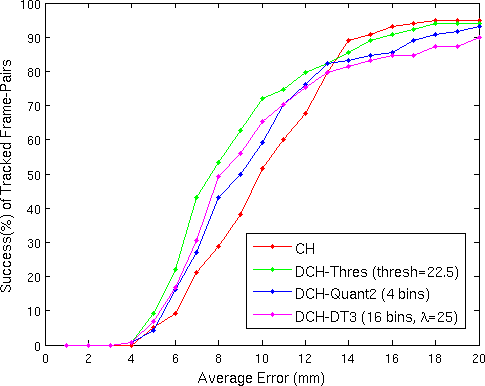}
				\label{fig:fig5_Subfig2}
			}
		\vspace{-2mm}
		\caption{
			Comparison of all distances with best settings.
			Although \textit{DCH-DT3} provides a smoother distance measure, \textit{DCH-Thres} performs best on both datasets.
		}
		\label{fig:fig5}
		\vspace{-5mm}
		\end{figure}
		
	    Fig.~\ref{fig:fig5} summarizes the results for each distance with the best parameter setting. 
	    As expected, the results show that directional information improves the estimation accuracy. 
	    However, it is not \textit{DCH-DT3} that performs best for hand pose estimation, but \textit{DCH-Thres}, which is also more efficient to compute.    
	    While for \textit{DCH-DT3} the full hand model converges smoothly to the final pose, the thresholding yields a better fit to the silhouette after convergence (\textit{see supplementary video}\footnote{\url{http://youtu.be/Cbu3eEcl1qk}}).
	    Comparing the performances between synthetic and real data, we conclude that synthetic data is a good performance indicator, but might be misleading sometimes. 
	    For instance, \textit{CH} performs well on the synthetic data but worst on the real data. 
	    This is also reflected by the mean error for the various frame differences provided in Table~\ref{table:errors}, that introduce an increasing difficulty in the benchmark. 
	    Denoted with the term \textit{initial} is the average 3d distance of the joints before running the pose estimation algorithm.
	    The result of a full tracking system \cite{LucaHands} is provided for comparison, which expectedly performs better due to the number of features combined.
	    Finally, runtime is provided for the synthetic experiments, giving some intuition about the time efficiency of each method.

	\begin{table}[t]	
		\vspace{-2mm}
	\footnotesize 
		\begin{center}
			\caption{
				Mean error$\pm$std.dev.(mm), av.time (sec) for 1,5,10,15 frame differences. Time measurements regard single-threaded code on a 6-core 3GHz Xeon PC.
			}
			\label{table:errors}
			\setlength{\tabcolsep}{1pt}	
			\begin{tabular}{|c|l|c|c|c|c|c| c|}
				\cline{3-8} 
				\multicolumn{1}{c}{} & \multicolumn{1}{c|}{} & $1$ & $5$ & $10$ & $15$ & All & Time\\
				\cline{3-8}		
				\noalign{\smallskip}
				\hline
				\multirow{4}{*}{\centering\begin{turn}{90}Synthetic\end{turn}} & {         \textit{CH}         } & { $1.0$$\pm$$1.0$ } & { $2.5$$\pm$$2.5$ } & { $4.3$$\pm$$4.6$ } & { $ 6.4$$\pm$$6.1$ } & {                  $3.5$$\pm$$4.5$ } & {$103$}\\\cline{2-8}
				                          									{} & {         \textit{DCH-DT3}    } & { $2.0$$\pm$$1.3$ } & { $2.3$$\pm$$1.3$ } & { $3.8$$\pm$$2.9$ } & { $ 6.2$$\pm$$5.8$ } & {                  $3.6$$\pm$$3.8$ } & {$115$}\\\cline{2-8}
				   															{} & {         \textit{DCH-Quant}  } & { $4.0$$\pm$$1.6$ } & { $4.2$$\pm$$1.7$ } & { $5.4$$\pm$$2.5$ } & { $ 7.0$$\pm$$4.0$ } & {                  $5.1$$\pm$$2.9$ } & {$161$}\\\hhline{|~|-------}
				       														{} & {         \textit{DCH-Thres}  } & { $1.1$$\pm$$0.8$ } & { $1.3$$\pm$$1.1$ } & { $2.5$$\pm$$2.4$ } & { $ 4.1$$\pm$$4.5$ } & { \cellcolor{Gray} $2.2$$\pm$$2.9$ } & {$077$}\\
				   
				\hline				
				\noalign{\smallskip}
				\hline
				
				\multirow{6}{*}{\centering\begin{turn}{90}Realistic\end{turn}} & {         \textit{Initial}                      } & { $ 6.4$$\pm$$ 2.0$ } & { $10.5$$\pm$$5.6$ } & { $16.5$$\pm$$11.5$ } & { $22.6$$\pm$$16.9$ } & {                  $14.0$$\pm$$12.3$ } & {-}\\\cline{2-8}
																		    {} & {         \textit{Ballan \etal~\cite{LucaHands}}} & { $ 5.9$$\pm$$ 1.9$ } & {         -        } & {         -         } & {         -         } & {                          -         } & {-}\\\cline{2-8}
																		    {} & {         \textit{CH}                           } & { $ 7.1$$\pm$$ 1.9$ } & { $ 7.8$$\pm$$2.4$ } & { $ 9.3$$\pm$$ 4.3$ } & { $10.9$$\pm$$ 5.9$ } & {                  $ 8.8$$\pm$$ 4.2$ } & {-}\\\cline{2-8}
																		    {} & {         \textit{DCH-DT3}                      } & { $ 6.3$$\pm$$ 1.5$ } & { $ 6.7$$\pm$$2.0$ } & { $ 8.7$$\pm$$ 5.1$ } & { $11.1$$\pm$$ 7.9$ } & {                  $ 8.3$$\pm$$ 5.4$ } & {-}\\\cline{2-8}
																	  		{} & {         \textit{DCH-Quant}                    } & { $ 6.8$$\pm$$ 1.6$ } & { $ 7.2$$\pm$$2.1$ } & { $ 9.0$$\pm$$ 4.4$ } & { $10.7$$\pm$$ 7.3$ } & {                  $ 8.4$$\pm$$ 4.7$ } & {-}\\\hhline{|~|-------}
																		    {} & {         \textit{DCH-Thres}                    } & { $ 6.1$$\pm$$ 1.3$ } & { $ 6.4$$\pm$$1.8$ } & { $ 7.6$$\pm$$ 3.3$ } & { $ 9.4$$\pm$$ 5.3$ } & { \cellcolor{Gray} $ 7.4$$\pm$$ 3.6$ } & {-}\\
				\hline
				\end{tabular}
		\end{center}
		\vspace{-7mm}
	\end{table}
		
\vspace{-1mm}
\section{Conclusion}\label{sec:conclusion}
\vspace{-3mm}
	In this work, we propose a new benchmark dataset for hand pose estimation that allows to evaluate single components of a hand tracker without running a full system.
	As an example, we discuss a generalized Chamfer distance and evaluate four special cases. 
	The experiments reveal that directional information is important and a signed circular distance performs better than an unsigned distance in the case of silhouettes. 
	Interestingly, a distance using a circular threshold outperforms a smooth directional Chamfer distance both in terms of accuracy and runtime.     
	We finally conclude that synthetic data can be a good indicator for the performance, but might be misleading when comparing different methods.
	Future plans include adding frame pairs of other sequences with more background clutter and segmentation noise.
	
\vspace{-2mm}

\section{Acknowledgments}\label{sec:acknowledgements}
\vspace{-3mm}
	The authors acknowledge financial support from the DFG Emmy Noether program (GA 1927/1-1) and the Max Planck Society.
	
\vspace{-3mm}

\bibliographystyle{splncs03}
\bibliography{egbib}

\end{document}